\documentclass{article}

\usepackage[nonatbib,final]{score_workshop}

\usepackage[utf8]{inputenc} 
\usepackage[T1]{fontenc}    
\usepackage{url}            
\usepackage{booktabs}       
\usepackage{amsfonts}       
\usepackage{nicefrac}       
\usepackage{microtype}      
\usepackage{xcolor}         
\usepackage{graphicx}
\usepackage{amsthm}
\usepackage{amsmath}
\usepackage{amssymb}
\usepackage{hyperref}       
\usepackage[capitalise]{cleveref}

\newtheorem{prop}{Proposition}

\newcommand{\majoru}{u_{+}}
\newcommand{\majorv}{v_{+}}
\newcommand{\majorw}{w_{+}}
\newcommand{\zerow}{w_{0}}
\newcommand{\minoru}{u_{-}}
\newcommand{\minorv}{v_{-}}
\newcommand{\minorw}{w_{-}}

\DeclareMathOperator{\Tr}{Tr}

\usepackage{wrapfig}
\newcommand{\thetitle}{Score-based generative models learn \\manifold-like structures with constrained mixing}
\title{\thetitle}

\author{%
Li Kevin Wenliang\\
DeepMind \\
\texttt{kevinliw@deepmind.com}\\
\And
Ben Moran\\
DeepMind \\
\texttt{benmoran@deepmind.com}\\
}

\usepackage[natbib=true,style=numeric,maxcitenames=1,uniquename=false,uniquelist=false]{biblatex}
\begin{document}

\maketitle

\begin{abstract}

How do score-based generative models (SBMs) learn the data distribution supported on a low-dimensional manifold? 
We investigate the score model of a trained SBM through its 
linear approximations and subspaces spanned by local feature vectors. 
During diffusion as the noise decreases, the local dimensionality increases and becomes 
more varied between different sample sequences. 
Importantly, we find that the learned vector field mixes samples by 
a non-conservative field within the manifold, although it denoises with normal projections 
as if there is an energy function in off-manifold directions. At each noise level, 
the subspace spanned by the local features overlap with an effective density function.
These observations suggest that SBMs can flexibly mix samples with the learned score field
while carefully maintaining a manifold-like structure of the data distribution.

\end{abstract}

\section{Introduction}

\begin{wrapfigure}[7]{r}{0.6\textwidth}
    \vspace{-1em}
    \includegraphics[width=0.59\textwidth]{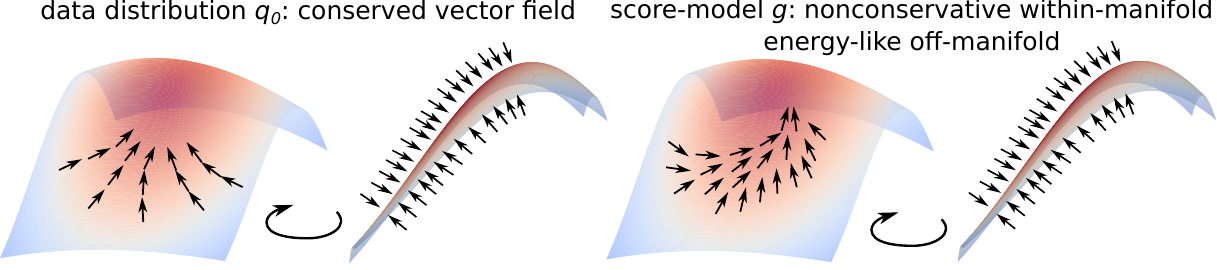}
    \caption{Schematic of our findings.}
    \label{fig:schematic}
\end{wrapfigure}


Score-based generative models (SBM) 
have achieved great success in learning natural data distributions \citep{song2019generative,song2020score}, which according to the manifold hypothesis are supported on low-dimensional manifolds.
Unlike decoder-based methods such as VAEs and GANs, 
most SBMs do not explicitly construct 
distributions on low-dimensional manifolds with a compressed representation. 
Then, how do SBMs approximate distributions on manifold?

Given a dataset $\{x^{(i)}\}$ drawn i.i.d.\ from an unknown data density 
$q_0$ on $\mathbb{R}^d$, a score model  
$g_{\sigma_t}(\cdot):\mathbb{R}^d\to \mathbb{R}^d$ 
estimates the score function of the noisy data distribution $\nabla_x \log q_{\sigma_t}(x)$, 
where $\sigma_t$ is the standard deviation of the additive Gaussian noise at time $t$, 
and $t\in[0,1]$.
Typically, as $t$ decreases, $\sigma_t$ decreases towards zero, 
and $q_1(x)$ is almost identical to a Gaussian.
The estimated score function can then be used to define 
a stochastic or deterministic process to
generate samples close to $q_0(x)$ 
by initialising from Gaussian samples.
This procedure forms the basic methodology of SBM.

\cref{fig:schematic} summarises our main finding. 
The left two panels are two rotated views of the example data density $q_0$ (blue-orange)
on a 2D manifold in a 3D ambient space.
The density implies a conserved vector field. 
The right two panels show the vector field of a 
score model $g$ which approximates the score of $q_0$
but is not guaranteed to be conservative.
We find that, in this score model, 
the vector field is non-conservative only within the manifold; 
whereas the field in directions normal to the manifold remains close to the 
conservative score field of the noisy data distribution, constraining the samples to stay around the data manifold. 
Further, the local features of $g$
span the same local subspace of an effective density function that is consistent with $g$ 
in the sense we clarify soon.

\section{Local orthogonal features of approximate score functions}
A local approximation of the score 
$g_\sigma(x) \approx g_\sigma(x_0) + \nabla_x g_\sigma(x)\vert_{x_0}(x-x_0)$
around $x_0$ involves the score Jacobian, so we use its singular value decomposition (SVD) to 
analyse $g_\sigma$ locally:
\begin{equation}
\nabla_x g_\sigma(x)=\sum_{i=1}^d u_i(x)s_i(x)v^\intercal_i(x),
\end{equation}
where the features $u_i(x),v_i(x)$ and $s_i(x)\ge0$ are, respectively, the $i$'th left singular vector, right singular vector, 
and singular value of $\nabla_x g_\sigma(x)$, ordered so that $s_i < s_j$ for $i<j$. 
If $g_\sigma$ is the score of a density function, 
then $g_\sigma$ is conservative, $\nabla_x g_\sigma(x)$ is symmetric (or normal), and $u_i=\pm v_i$ for all $i$. 
To build more intuitions about score Jacobians, 
consider the multivariate Gaussian $\mathcal{N}(x; \mu, \Sigma)$ which has a constant score Jacobian $-\Sigma^{-1}$. 
Each pair of its singular vectors have \emph{opposite signs}. 
In particular, the \emph{singular value} of rank $i$ is equal to the 
\emph{inverse variance} along the $i$'th singular vector. 

Now, suppose that the data distribution $q_0$ is a low-dimensional Gaussian supported on a subspace of $\mathbb{R}^d$.
After adding a small additive Gaussian noise on $\mathbb{R}^d$, the SVD of the score Jacobian shows interpretable properties of the data distribution: 
large singular values or small variances appear along
directions with abrupt changes in the score, 
reflecting steep curvatures along the off-manifold 
directions. Conversely, small singular values or large variances are associated with
on-manifold directions along which the data density varies smoothly. This pattern
generalises to curved manifolds as long as the noise is small compared to the local curvature.
In practice, the Jacobian of a learned score estimator $\nabla_x g_\sigma(x)$ may not be 
symmetric as that of the Gaussian, but we can compare it to
an effective conservative (energy-based) score field.
\begin{prop}
Given a model score $g_{\sigma_t}(x)$, there exists a conservative field $\tilde{g}_{\sigma_t}(x)$
such that for every path $x_{1:t}$ from $g_{\sigma_t}$, 
the likelihood of $x_t$ computed under $g_{\sigma_t}$ and $\tilde{g}_{\sigma_t}$ are identical.
\end{prop}
\vspace{-1.4em}
\begin{proof}
Suppressing the subscripts for brevity, we decompose $\nabla{g}$ as the sum of 
a symmetric component $\nabla \tilde{g}:=0.5(\nabla g + \nabla^\intercal \! g)$ and 
a skew-symmetric component $H:=0.5(\nabla g - \nabla^\intercal\! g)$. 
By Poincar\'e's Lemma, the symmetric $\nabla\tilde{g}$
implies the existence of a conservative field $\tilde{g}$. 
Given a sample path $x_{1:t}$, the likelihood of $x_t$ depends on $g$ 
through its divergence or the trace of the score Jacobian 
$\Tr(\nabla \tilde{g})$ \citep[eqn. 39]{song2019generative}.
Then, we have $\Tr(\nabla g) =  \Tr(\nabla \tilde{g} + H) = \Tr(\nabla \tilde{g}) + \Tr(H)  = \Tr(\nabla \tilde{g}) + 0  = \Tr(\nabla \tilde{g})$.
\end{proof}
\vspace{-0.8em}
This means that, \emph{given a sample path}, $\tilde{g}_\sigma$ is a valid score of 
an (unknown) 
density function equivalent to $g_\sigma$ in terms of $x_t$'s likelihood. 
Similar to but unlike the SVD of $\nabla_x{g}_\sigma(x)$, 
the eigendecomposition of the symmetric $\nabla_x\tilde{g}_\sigma(x)$ reveals the 
local features of the equivalent density: 
following the intuition of the Gaussian distributions, 
we see that the eigenvectors with negatively large 
eigenvalues correspond to off-manifold directions ($-\Sigma^{-1}$ is negative semi-definite); eigenvectors with small-in-magnitude eigenvalues 
indicate on-manifold directions. Positively large eigenvalues indicate positive curvature, and we find that 
they exhibit on-manifold features as shown in our experiments.

\section{Experiments}

We now empirically compare the SVD of $\nabla_x g_\sigma(x)$ and the eigendecomposition of $\nabla_x \tilde{g}_\sigma(x)$ 
to discover local geometries of a learned score model. 
For ease of visualisation, 
we used an SBM trained on MNIST digits \citep{lecun1998gradient}, so $d=28 \times 28=784$. The model is based on a U-net, 
adapted from the model of \citet{song2020score}. More details are in \cref{sec:model_details}.
For a given generated $x$ at noise level $\sigma$, 
we denote the (minor) left and right singular vectors corresponding to the 
smallest singular value of $\nabla_x g_\sigma(x)$ as 
$\minoru$ and $\minorv$, respectively, 
and the (major) vectors to the largest singular value as $\majoru$ and $\majorv$. 
For the ``symmetrised'' score $\nabla_x \tilde{g}_\sigma(x)$, we denote the minor and major eigenvectors as $w_-$ and $w_+$, respectively, and the eigenvector with smallest-in-magnitude eigenvalue as $w_0$.

\subsection{Local principal directions}

\begin{figure}
    \centering
    \includegraphics[width=0.8\textwidth]{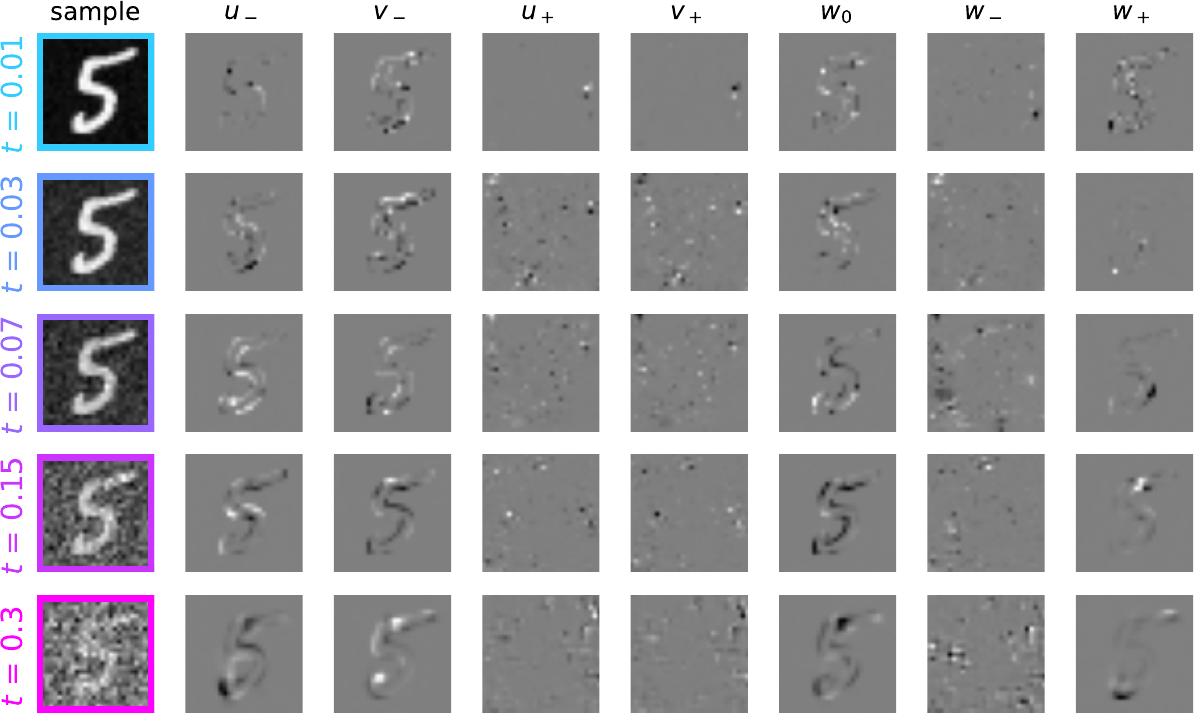}
    \caption{An example image sequence generated from the SBM and the singular and eigen-vectors.}
    \label{fig:img_svd}
\end{figure}
\cref{fig:img_svd} shows the generated samples at five time steps in the first column 
and their feature vectors in the other columns. 
Vectors $\minorv$, $\minoru$ and $\zerow$ reflect more general distortions of the strokes at higher noise levels, 
and more localised changes at lower noise levels.
However, $\minoru$ and $\minorv$ are not well aligned, 
reflecting non-conservation of the score model, or a non-normal score Jacobian.
The eigenvectors $\zerow$ and $\majorw$ of the symmetrised Jacobian
show similar on-manifold semantics, but they are also not aligned with the singular vectors.
Major vector $\majoru$ and $\majorv$
show high-frequency contents in the background, 
which are likely off-manifold directions, as following these directions makes
a sample less realistic. 
Unlike the minor singular vectors, these major vectors differ by a sign and points towards opposite directions.
The eigenvector $\minorw$ of the effective 
energy-based field also shows high-frequency, off-manifold patterns in the background, 
similar to the major singular vectors.

\begin{figure}
    \centering
    \includegraphics[width=0.495\textwidth]{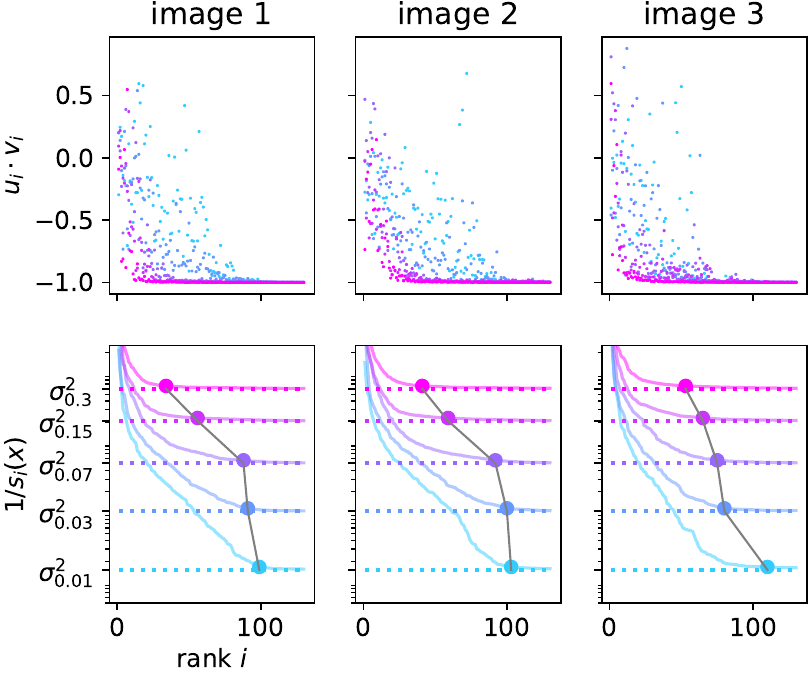}
    \includegraphics[width=0.495\textwidth]{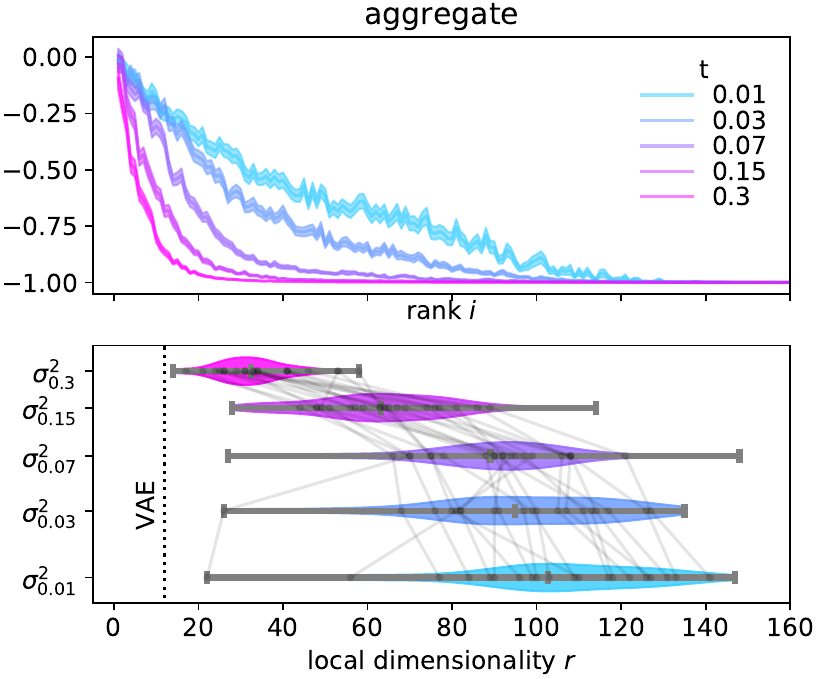}
    \caption{Top, cosine angle between the singular vectors. Bottom, singular values spectrum and effective 
    local dimensionality corresponding to singular values above the noise floor.}
    \label{fig:svd}
\end{figure}

In \cref{fig:svd} (top), we show the inner products $u_i(x) \cdot v_i(x)$ (cosine angle) for 
three other example image sequences and the summary statistics of 128 images. 
They confirm that the singular vectors are not aligned for lower ranks 
and they differ by a sign for higher ranks. 
The opposite signs between $u_i$'s and $v_i$'s in higher ranks
are consistent with the Gaussian score along off-manifold directions. 
But how to interpret the misalignment at lower ranks? 
The example image at $t=0.3$ in \cref{fig:img_svd} gives an interesting clue: 
its $\minorv$ focuses on the outline of  
the digit 5 but then maps to a $\minoru$ that resembles the outline of 6.
This misalignment of singular vectors mixes data 
across different components, and we speculate that it could facilitate 
mixing between high-level image classes at high noise levels. 

\begin{figure}
    \centering
    \includegraphics[width=0.99\textwidth]{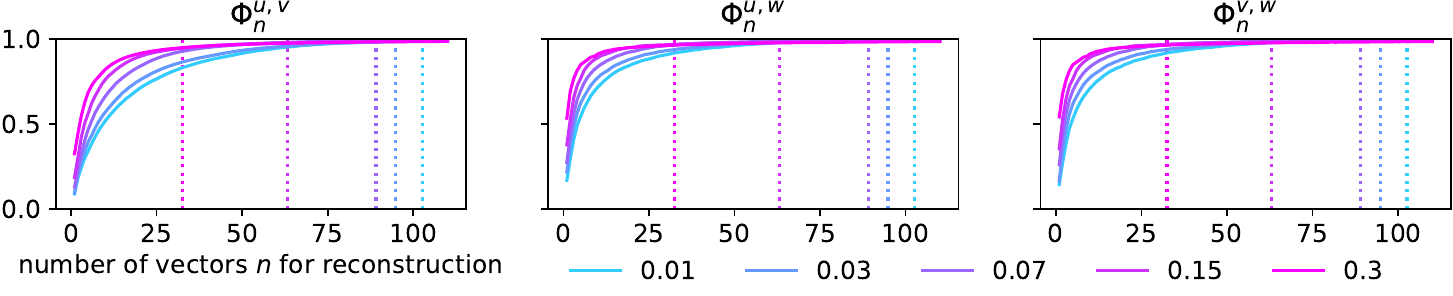}
    \caption{Solid lines: the amount of overlap between local subspaces. Dotted lines: mean 
    number of nontrivial ranks for each temperature from \cref{fig:svd} (bottom right).}
    \label{fig:overlap}
\end{figure}

\subsection{Singular values reveal a progressive and diverse manifold expansion}
Following from our Gaussian intuition, 
the inverse singular values of a score Jacobian are related to the local length scales along 
principal directions of the distribution. 
We should then expect to see inverse singular values equal to the corresponding 
noise variances $\sigma_t^2$ along many potentially off-manifold directions.
The examples in \cref{fig:svd} (bottom left) support this prediction. 
In addition, the noise variance $\sigma_t^2$ lower bounds the inverse singular values, 
because the smoothing effect of the Gaussian noise determines the smallest length scale. 

All nontrivial inverse singular vectors above the noise levels contribute to local data dimensionality, 
so we estimated the local
data dimensionality of $x_t$ by
the smallest $r_t$ such that $s^{-1}_i(x_t)<(1+\kappa)\sigma^2_t$ for all $i>r_t$, 
where we set $\kappa=0.1$ as a slack constant. 
The coloured dots in \cref{fig:svd} (bottom left) show the 
estimated local dimensionality for three images at each $t$, and \cref{fig:svd} (bottom right)
shows the distribution of thus estimated local dimensionalities for 128 images: 
they increases and spreads out more with decreasing noise levels.
We compared this with
the local dimensionality estimated by a VAE model \citep{dai2018diagnosing,dai2021value}, 
described in \cref{sec:vae_details}.
The VAE-estimated local dimensionality of clean images turned out to be within $13\pm3$, substantially lower and 
less varied than the estimates from the score model in \cref{fig:svd}. 
Thus, these patterns of singular values suggest that the SBM has a 
much more expressive and more flexible representation power
than the VAE.

\subsection{Misaligned singular vectors span the same subspace and provides non-normal mixing.}

From \cref{fig:img_svd,fig:svd}, we see that the singular vectors and eigenvectors are misaligned in the 
subspace given by nontrivial singular values. 
In principle, the right singular vectors define an input 
space of local changes, and the left singular vectors 
define an output basis to apply the denoising vector.
To at least preserve mixing during diffusion, the left and right subspaces 
spanned by nontrivial ${u_i}$ and ${v_i}$ must substantially overlap, since otherwise the score field
can project data into off-manifold directions. Further, they should also overlap with 
the eigen-subspace of the effective energy-based density spanned by 
$\{w_i\}$ with small-in-magnitude eigenvalues.
We quantify the amount of overlap between these subspaces 
by how well one reconstructs the other for the first $n$ vectors:
\begin{equation}
    \Phi_n^{a,b} := 1-\frac{1}{n}\sum_{j=1}^n \|\hat{a}_{j}^{b,n} - a_j\|_2^2 /\|a_j\|_2^2, ~\mathrm{where}~ 
    \hat{a}_j^{b,n} = \sum_{k=1}^n (b_k\cdot a_j)b_k, ~~a,b\in\{u,v,w\}.
\end{equation}
This is a criterion based on explained variance---a larger value indicates a better overlap.
The results in \cref{fig:overlap} confirm a considerable overlap between the three 
subspaces within the nontrivial ranks at each $t$ (\cref{fig:svd}). 
Therefore, although $g_\sigma(x)$ mixes image features onto different components
in a non-normal fashion, 
it still updates the image within highly overlapped subspaces. 
The singular subspaces of $\nabla_x g_\sigma(x)$ also overlap well with 
the eigen-subspace of the symmetrised $\nabla_x\hat{g}_\sigma(x)$, suggesting that 
the learned singular subspaces conform to the data subspace of the effective density. 

\section{Conclusion}
We have shown that the score field of a trained score-based model shows low local dimensionality consistent with the manifold hypothesis.
As the noise scale decreases, the modelled manifold dimensionality rises and becomes more varied, 
reflecting increasingly precise description of samples.
In addition to projecting noisy, off-manifold 
components back onto the manifold by normal projections, 
the score function mixes between feature directions within
the data manifold using a non-conservative field, giving 
misaligned singular vectors. 
Nonetheless, these vectors span overlapping input and output subspaces, 
both of which also overlap with the local eigen-subspace of the 
effective energy-based field.
Therefore, the score-based generative model learns a vector field that carefully 
constrains the samples to be within the manifold 
while mixing samples within the manifold. 
This explains the well-known harmless effect of non-conservation \citep{salimans2021should} 
and the little improvement obtained by encouraging conservation \citep{chao2022quasi}.
The discussion of 
a related work by \citet{mohan2019robust} is in \cref{sec:related}.

\paragraph*{Acknowledgement}
We thank Conor Durkan, Mikołaj Bińkowski, Wendy Shang and Sander Dieleman for
sharing their implementation of the diffusion model.


\printbibliography

@inproceedings{mohan2019robust,
  title={Robust And Interpretable Blind Image Denoising Via Bias-Free Convolutional Neural Networks},
  author={Mohan, Sreyas and Kadkhodaie, Zahra and Simoncelli, Eero P and Fernandez-Granda, Carlos},
  booktitle={International Conference on Learning Representations},
  year={2019}
}

@inproceedings{dai2018diagnosing,
  title={Diagnosing and Enhancing VAE Models},
  author={Dai, Bin and Wipf, David},
  booktitle={International Conference on Learning Representations},
  year={2018}
}

@article{dai2021value,
  title={On the Value of Infinite Gradients in Variational Autoencoder Models},
  author={Dai, Bin and Wenliang, Li and Wipf, David},
  journal={Advances in Neural Information Processing Systems},
  year={2021}
}

@inproceedings{song2020score,
  title={Score-Based Generative Modeling through Stochastic Differential Equations},
  author={Song, Yang and Sohl-Dickstein, Jascha and Kingma, Diederik P and Kumar, Abhishek and Ermon, Stefano and Poole, Ben},
  booktitle={International Conference on Learning Representations},
  year={2020}
}

@article{song2019generative,
  title={Generative modeling by estimating gradients of the data distribution},
  author={Song, Yang and Ermon, Stefano},
  journal={Advances in Neural Information Processing Systems},
  volume={32},
  year={2019}
}

@article{lecun1998gradient,
  title={Gradient-based learning applied to document recognition},
  author={LeCun, Yann and Bottou, L{\'e}on and Bengio, Yoshua and Haffner, Patrick},
  journal={Proceedings of the IEEE},
  year={1998},
}

@inproceedings{salimans2021should,
  title={Should EBMs model the energy or the score?},
  author={Salimans, Tim and Ho, Jonathan},
  booktitle={Energy Based Models Workshop-ICLR 2021},
  year={2021}
}

@article{chao2022quasi,
  title={Quasi-Conservative Score-based Generative Models},
  author={Chao, Chen-Hao and Sun, Wei-Fang and Cheng, Bo-Wun and Lee, Chun-Yi},
  journal={arXiv preprint},
  year={2022}
}

\clearpage

\appendix

\begin{center}
    \Large
    \textbf{\thetitle: \\ appendix}
\end{center}

\section{The score-based model}\label{sec:model_details}

The model, training objectives and procedures are 
essentially the same as the one proposed by \citep{song2020score}. 

\subsection{Denoising networks}

We took a standard implementation of the Unet and reduces the input size to $28 \times 28 \times 1$. 
The batch size is 2\,048 and the learning rate is $0.0003$. We trained the model for 540\,000 steps.

\subsection{Diffusion process}
After obtaining the score estimates, we generated the images by discretising the SDE
by the Euler-Maruyama method.
The noise schedule is a monotically decreasing function of $t$, and we apply appropriate scaling 
on the drift term of the SDE (variance-preserving). We took samples at five noise levels at 
$t\in\{0.01,0.03, 0.05, 0.15,0.3\}$, approximately corresponding to noises with standard deviations 
$\sigma_t\in\{0.039, 0.10, 0.22, 0.46, 0.77\}$. For all summary statistics we used 128 randomly sampled
images from the model so that they stay in the learned but approximate score field. Using more images does not change 
the conclusions.

\section{VAE model}\label{sec:vae_details}

Following \citep{dai2018diagnosing, dai2021value}, we estimated the local dimensionality 
of a data point by the number of latent dimensions in which the posterior variances 
were close to zero. 
In practice, we regarded any posterior variance less than 0.5 as being close to zero. 
The VAE model we used to obtain a baseline data dimensionality has a standard Gaussian prior on $\mathbb{R}^{100}$. 
The likelihood is a Gaussian with its mean given by a neural network function of the latent variable and with variance as 
a free positive parameter. 
The variational posterior is a factorised Gaussian
with means and variances as output of a neural network function. 
All neural networks have convolutional, ReLU nonlinearity and batch normalisation layers similar to the DCGAN.

We trained the network for 1\,000 epochs on 60\,000 MNIST digits. The estimated local dimensionality
varies between digits: the digit ``1'' in general has lower dimensionality (around 11) than the other digit 
(around 18). This number decreases when we trained on noisy images, consistent with the effect of noise smoothing.

\section{Related work}\label{sec:related}

\citet{mohan2019robust} used SVD to analyse locally linear features of the trained denoising model 
which was not used for diffusion-based generation.
There, the trained model $f(\cdot):\mathbb{R}^d\to\mathbb{R}^d$ outputs the denoised image 
under the expected $l$-2 loss directly rather than 
the score function. Also, the networks are 
constrained to be bias-free and have ReLU nonlinearities, 
so that the network is locally affine: $f(x)=\nabla f(x)x$. 

We kept our score network to be as general as possible so did not impose these constraints. 
If we adopt these constraints, we will have $g(x)=\nabla g(x)x$. 
Under Gaussian noise with variance $\sigma^2$, when both $f$ and $g$ are perfectly trained, the optimally (in terms of expected $l$-2 loss) denoised estimate is 
\begin{equation}\label{sec:bias_free}
    \hat{x}=x + \sigma^2 g(x)=x+\sigma^2\nabla g(x) x = (\mathbf{I} + \sigma^2 \nabla g(x)) x = \nabla f(x)x,
\end{equation}
a form of the Stein's unbiased risk estimate. Clearly, the $f$ used by \citet{mohan2019robust} and the $g$ used by us 
are equivalent parametrisations of each other.

One should then expect to find the same conclusions between this paper and theirs. 
A seemingly contradictory finding is that, in \citet{mohan2019robust}, the left and right singular vector of $\nabla f(x)$ at a given rank 
were mostly aligned , whereas those of $\nabla g(x)$ here are mostly opposite to each other.
This is, however, expected. In the idealised case where the score function 
is perfectly estimated, $\nabla g$ is symmetric, and so is $\nabla f$. For small noise, the score Jacobian 
$\nabla g$ is symmetric with eigenvalues lower-bounded by $-1/\sigma^2$, since this is the lowest noise
present in the data. Applying the tranformation between $f$ and $g$ in \eqref{sec:bias_free} 
shows that $\nabla f$ is in fact positive semidefinite, which has identical left and right singular vectors. 
However, \citet{mohan2019robust} did not interpret the misaligned singular vectors, which we explain 
as arising from the non-conservation of the approximate score model.


\end{document}